\documentclass{article}

\usepackage{microtype}
\usepackage{graphicx}
\usepackage{subcaption}
\usepackage{booktabs} %

\usepackage{hyperref}
\usepackage{tikz}
\usetikzlibrary{arrows.meta, positioning, calc}
\usepackage[dvipsnames]{xcolor}

\usepackage[preprint]{icml2026}

\usepackage{amsmath}
\usepackage{amssymb}
\usepackage{mathtools}
\usepackage{amsthm}
\usepackage{comment}
\usepackage{tabularx}
\usepackage{pgfplots}
\usetikzlibrary{calc}

\usepackage[capitalize,noabbrev]{cleveref}

\theoremstyle{plain}

\theoremstyle{definition}

\theoremstyle{remark}

\usepackage[textsize=tiny]{todonotes}

\icmltitlerunning{Large Language Models Should Learn Personalized Rather Than Aggregated Human Preferences}

\begin{document}

\twocolumn[
  \icmltitle{Large Language Models Should Learn Personalized Rather Than Aggregated Human Preferences}

  \icmlsetsymbol{equal}{*}

  \begin{icmlauthorlist}
    \icmlauthor{Cristina G\^arbacea}{yyy}
  \end{icmlauthorlist}

  \icmlaffiliation{yyy}{University of Chicago, Data Science Institute, Chicago, USA}

  \icmlcorrespondingauthor{}{garbacea@uchicago.edu}

  \icmlkeywords{Machine Learning, ICML}

  \vskip 0.3in
]

\printAffiliationsAndNotice{}  %

\begin{abstract}
Current approaches to aligning large language models (LLMs) 
aggregate diverse human preferences into a single reward signal, 
effectively optimizing for a hypothetical ``average user'' who 
represents no real person particularly well. This position 
paper argues that LLMs should learn personalized, individual 
preferences rather than aggregated ones. We show that aggregation 
masks critical information about preference diversity, individual 
values, and contextual dependencies, which is a limitation both 
theoretically grounded in social choice theory and empirically 
evident across demographic groups. We analyze the rich structure 
that human preferences encode, survey technical approaches to 
personalization, and systematically address counterarguments on 
scalability, shared standards, and manipulation risk. While 
personalization introduces genuine safety challenges including 
filter bubbles, value lock-in, and psychological manipulation, we argue these are manageable through bounded personalization 
frameworks that preserve universal safety constraints while accommodating legitimate individual variation. We conclude with a concrete research and policy agenda for developing 
preference-aware models that respect both individual autonomy 
and collective safety.
\end{abstract}

\section{Introduction}

The dominant paradigm for aligning large language models with human values relies on learning from preference data~\citep{ouyang2022training,bai2022training}. In reinforcement learning from human feedback (RLHF), models are fine-tuned using a reward model trained on human preference comparisons, typically of the form ``output A is preferred to output B for prompt P''. This approach has produced remarkable improvements in model helpfulness, harmlessness, and honesty~\citep{askell2021general}. However, this paradigm rests on a problematic assumption: that human preferences can be meaningfully aggregated into a single reward signal. In practice, annotators exhibit substantial disagreement~\citep{gordon2022jury,aroyo2015truth}, preferences vary across cultural contexts~\citep{huang2023culturally, santurkar2023whose}, and individual users have distinct values and needs~\citep{kirk2023personalisation}. Training on averaged 
preferences thus optimizes for a hypothetical ``average user'' 
who may not actually exist and whose preferences match no real person particularly well. The problem compounds with 
scale: models trained on preference annotations from a few thousand 
participants are deployed to billions of 
users across hundreds of languages, cultures, and contexts whose 
preferences were never represented in training.

The costs of this design choice are not abstract. When asked to explain a technical concept, some users prefer concise definitions while others want detailed explanations with examples; some value formal mathematical notation while others prefer intuitive analogies. These are not disagreements about quality but legitimate differences in 
what constitutes a helpful response for a given person in a given context. Yet current training approaches collapse these differences into a single averaged reward signal, producing responses that partially satisfy everyone while fully satisfying no one.

This paper argues that we must move beyond aggregated preference training toward personalized and adaptive systems that respect the diversity of human preferences.
We examine the 
theoretical and empirical limitations of aggregation, analyze the 
rich multidimensional structure of human preferences, and make the 
case for personalization as both beneficial and necessary for serving heterogeneous user populations. We critically examine counterarguments including scalability, shared standards, 
manipulation risk, and propose a responsible personalization 
framework that preserves universal safety constraints while 
accommodating legitimate individual variation. As language models become infrastructure for information access and decision-making at an increasingly global scale, AI alignment research must reckon with a fundamental question: \textit{aligned for whom?}

\section{Limitations of Aggregated Preferences}

\subsection{The Impossibility of Universal Preferences}

\pgfplotsset{compat=1.18}

\begin{figure}[t]
\centering
\begin{tikzpicture}
\begin{axis}[
  width=0.9\columnwidth, height=6.5cm,
  xlabel={Communication style (informal $\to$ formal)},
  ylabel={Technical depth (accessible $\to$ technical)},
  xtick=\empty,
  ytick=\empty,
  xmin=0, xmax=10, ymin=0, ymax=10,
  axis line style={-stealth},
  label style={font=\small},
  legend cell align={left},
  legend style={at={(0.98,0.02)}, anchor=south east,
                font=\scriptsize},
  clip=false
]

\addplot[only marks, mark=*, mark size=6pt,
  draw=teal!80!black, fill=teal!70, opacity=0.85,
  legend image post style={mark size=4pt}]
  coordinates {(7.8,8.0)(8.2,8.4)(7.5,7.6)
               (8.4,7.9)(8.0,8.7)(7.7,7.3)(8.6,8.1)};
\addlegendentry{Technical, formal users}
\draw[teal!70!black, dashed, thick]
  (axis cs:8.0,8.0) ellipse [x radius=1.4, y radius=1.5];

\addplot[only marks, mark=*, mark size=6pt,
  draw=orange!90!red, fill=orange!70, opacity=0.85,
  legend image post style={mark size=4pt}]
  coordinates {(1.8,1.6)(2.2,2.0)(1.5,1.2)
               (2.4,1.7)(1.7,2.2)(2.0,0.9)(1.3,1.8)};
\addlegendentry{Accessible, informal users}
\draw[orange!70!red, dashed, thick]
  (axis cs:1.9,1.6) ellipse [x radius=1.4, y radius=1.5];

\addplot[only marks, mark=*, mark size=6pt,
  draw=violet!80, fill=violet!60, opacity=0.85,
  legend image post style={mark size=4pt}]
  coordinates {(1.4,7.8)(1.8,8.2)(1.1,7.4)
               (2.0,7.9)(1.5,8.5)(1.7,7.2)(1.2,7.7)};
\addlegendentry{Technical, informal users}
\draw[violet!60, dashed, thick]
  (axis cs:1.5,7.8) ellipse [x radius=1.3, y radius=1.5];

\addplot[only marks, mark=*, mark size=8pt,
  draw=gray!70, fill=gray!55, opacity=0.9,
  legend image post style={mark size=4pt}]
  coordinates {(4.5,5.2)};
\addlegendentry{Aggregated reward model}
\node[font=\scriptsize, gray, anchor=west]
  at (axis cs:4.8,5.2) {Aggregated reward model};
\node[font=\scriptsize, gray, anchor=west]
  at (axis cs:4.8,4.6) {\textit{(no real user is here)}};

\end{axis}
\end{tikzpicture}
\caption{Diverse user preferences cluster in distinct regions of 
preference space. The aggregated reward model (gray) falls in a 
sparse region between clusters, representing no actual user group 
and systematically failing minority populations.}
\label{fig:preference-space}
\end{figure}
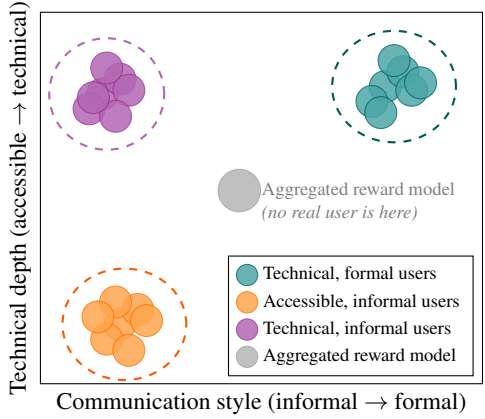

Aggregating diverse human preferences into a single reward signal 
is not merely impractical, it is theoretically impossible to do 
without imposing value choices. Arrow's impossibility theorem~\citep{arrow1950difficulty} demonstrates 
that no aggregation method can simultaneously satisfy transitivity, 
non-dictatorship, Pareto efficiency, and independence of irrelevant 
alternatives when combining individual preferences into collective decisions. 
We invoke this as a conceptual motivation: 
while the formal conditions do not transfer 
directly (Arrow concerns ordinal rankings 
over discrete alternatives, whereas RLHF optimizes continuous reward 
functions via probabilistic aggregation), the core insight does -- no aggregation of 
heterogeneous preferences is value-neutral. Human preferences 
exhibit systematic intransitivity and context-dependence when 
aggregated across diverse annotators~\citep{tversky1969intransitivity, 
azar2024general}, and analogous impossibility results have been 
identified in computational social choice 
settings~\citep{conitzer2024position, noothigattu2018voting}. Current RLHF implementations sidestep this impossibility by simply averaging preferences; disagreements are treated as noise rather than meaningful 
signal, a pragmatic choice with systematic consequences for who ultimately 
gets served well by deployed models (Figure~\ref{fig:preference-space}).

 When preferences are aggregated through majority voting or averaging, minority viewpoints are suppressed. If 60\% of annotators prefer direct responses while 40\% prefer nuanced caveats, models trained on averaged preferences will optimize for directness, making them poorly suited for users who value nuance. This is not merely a matter of slightly reduced performance; for users whose preferences consistently diverge from the majority, the model may be fundamentally unsuitable for their needs. The problem compounds across multiple preference dimensions: a user whose preferences 
fall in the minority on formality, detail level, and communication 
style simultaneously receives a model that is misaligned on all these dimensions at once. 

The loss of minority perspectives is particularly concerning given the demographics of preference annotation. Current datasets predominantly represent English-speaking, Western, educated populations~\citep{kirk2023personalisation, mihalcea2025ai}, embedding 
particular cultural norms while marginalizing alternatives. 
LLM opinions correlate up to 0.3 points higher with liberal, educated, 
Western populations than with other demographic groups~\citep{santurkar2023whose}. Global South perspectives are 
systematically under-represented, with alignment scores 
consistently lower than those observed for Western nations~\citep{durmus2024towards}. 
When prompted with culturally variable questions such as ``What does a 
wedding look like?'', models default to Western conventions, 
largely ignoring the diversity of cultural practices worldwide~\citep{nakano2021webgpt}. Models further systematically mispredict minority annotator preferences 
even when majority preferences are well-captured, with annotation 
disagreement reflecting genuine preference diversity rather than 
labeling noise~\citep{fleisig2023majority, jiang2025survey}.

Beyond demographic heterogeneity, preferences are highly 
context-dependent~\citep{shen2023large}, varying with task type, 
user expertise, and situational factors. A software engineer 
debugging production code wants different explanations than a 
student learning programming concepts for the first time; a medical 
professional requires different health information than a patient 
seeking to understand their diagnosis; a user under time pressure 
needs different response characteristics than one casually exploring 
ideas. Aggregated training washes out these contextual dependencies, 
treating all requests as equivalent regardless of their context.

\subsection{Misalignment Between Training and Deployment}

The impossibility of principled preference aggregation creates a 
deeper structural mismatch: we train a single reward model on 
preferences from a limited annotator pool, then deploy it to serve 
billions of users across diverse cultural, linguistic, and social 
contexts. This approach forces decisions using a reward signal that 
is suboptimal for nearly everyone -- a phenomenon we term 
``preference mediocrity.'' Like designing a single shoe size for 
all humans, it fits no one well while claiming to serve everyone.

The temporal dimension of this mismatch compounds the problem. 
Preferences are learned from fixed datasets collected at particular 
points in time, but must generalize to evolving user needs and 
contexts. A user's preferences for formality, detail level, or 
risk tolerance may shift with their expertise, urgency, or emotional 
state~\citep{jakesch2023co}. A novice user's preferences evolve as 
they gain expertise; a user's tolerance for speculation versus 
certainty varies between casual exploration and critical 
decision-making. Static reward models cannot capture these dynamics.

By optimizing for aggregated preferences, we implicitly encode particular cultural norms, communication styles, and value systems into model behavior~\citep{leon2025principles}. Models become vehicles for these dominant preferences, shaping how billions of users interact with information and ideas, with consequences that
extend beyond individual experience to epistemic practices, social 
communication, and the diversity of perspectives in public discourse~\citep{jiang2026artificial}.

\subsection{Technical Limitations}

These conceptual problems are compounded by technical challenges 
inherent to aggregated preference training. Reward models trained 
on preference data exhibit high 
uncertainty~\citep{sun2025rethinking, sun2024rethinking}, 
particularly for inputs far from the training distribution. 
Aggregation exacerbates this by forcing the reward model to average 
over diverse and potentially contradictory signals, making it 
fundamentally uncertain about what it should optimize for: it cannot 
distinguish cases where all annotators agree from cases where 
preferences are evenly split.

This uncertainty creates exploitable surfaces for reward hacking~\citep{wang2026reward}. 
Models learn to exploit imperfections in reward models through 
gaming strategies~\citep{gao2023scaling}, generating responses 
that score highly on the reward model while failing to genuinely 
satisfy user preferences. Aggregated reward surfaces are 
particularly vulnerable because averaging over inconsistent signals 
creates spurious correlations/artifacts that models learn 
to exploit.

Finally, truly comprehensive preference coverage is economically 
infeasible under current paradigms. Even well-resourced projects 
collect at most hundreds of thousands of preference 
judgments~\citep{ouyang2022training}, which is a tiny fraction of what 
would be needed to capture meaningful variation across billions 
of users and contexts. We discuss concrete strategies for 
overcoming both computational and data collection hurdles in 
Section~\ref{sec::scalability}.

\section{What Do Preferences Encode?}

Understanding what human preferences capture is essential for designing better alignment approaches. Preferences are not monolithic judgments but complex signals encoding multiple interacting factors, from task type and user expertise to cultural context and individual 
values. Appreciating this multidimensional 
structure is not merely descriptive: it reveals precisely what 
aggregation discards and what personalized systems must recover. We examine what preferences encode and how that structure can be discovered.

\subsection{Factors Underlying Preference Judgments}

Preferences vary significantly with task type. For summarization, 
users may prioritize brevity and coverage; for creative writing, 
elaboration and stylistic flair; for factual questions, accuracy and 
clarity; for open-ended discussion, engagement and 
perspective-taking~\citep{zheng2023judging}. A preference for 
directness in one context may thus coexist with a preference for 
circumspection in another, making any single aggregated signal 
ill-suited to the diversity of tasks.

User expertise and background profoundly shape preferences. Expert 
users favor technical depth, precise terminology, and minimal 
explanation of background concepts; novices prioritize accessibility 
and connection to familiar ideas. The same explanation can be 
perfectly calibrated for one user while incomprehensible or 
condescending to another. These differences reflect not only 
stylistic preferences, but fundamental differences in what represents helpful and useful information to different users.

Underlying all preferences are individual values and beliefs. 
Preferences regarding political ideology, risk tolerance, privacy 
concerns, and moral frameworks cannot be 
reconciled through averaging without imposing particular value 
systems~\citep{sorensen2024roadmap}. When a user prefers responses that acknowledge uncertainty, this may reflect epistemic humility as a value. When another prefers confident assertions, this may reflect a different relationship to authority and expertise. Both are legitimate perspectives grounded in different worldviews, yet aggregation forces models to stake out a middle position that may satisfy neither.

Cultural and linguistic context introduces further variation. 
Communication norms, politeness conventions, and acceptable content 
vary dramatically across cultures~\citep{huang2023culturally}. 
Preferences learned primarily from English-speaking, Western 
annotators embed particular cultural norms, potentially creating 
awkward or inappropriate interactions for users from other 
backgrounds.

Finally, situational factors modulate preferences even for 
individual users. Urgency, stress, confidence, and goals all affect 
what responses are appropriate~\citep{jakesch2023co}. The same user 
may prefer different responses depending on whether they are casually 
exploring a topic or making a critical decision. These contextual 
variations cannot be captured by any static preference model.

\subsection{Discovering Preference Structure}

How can we uncover the latent structure in preference data? 
Several approaches show promise for revealing preference structure 
by treating annotator disagreement as signal rather than noise. 
Modeling preference data as arising from multiple latent user 
types~\citep{jacobs1991adaptive, gordon2022jury} identifies distinct 
preference clusters, while multidimensional reward 
representations~\citep{wu2023fine} capture fine-grained trade-offs 
across helpfulness, harmlessness, truthfulness, verbosity, and 
technical depth. Causal 
inference~\citep{park2024disentangling} identifies which factors 
(expertise, context, demographics) drive preference judgments, 
enabling more robust generalization than correlation alone. 
Interpretable reward models~\citep{nanda2023progress, reber2025rate} 
allow inspection of what features drive preference predictions, 
supporting debugging and validation. Contrastive analysis across 
demographic groups and tasks~\citep{santurkar2023whose} reveals where 
aggregation is most harmful, guiding personalization efforts toward 
the dimensions that matter most. Finally, active preference 
elicitation~\citep{sadigh2017active, li2025eliciting} can rapidly 
identify a user's position in preference space from strategically 
chosen queries, and preference learning from demonstrations creates more 
personalized and effective 
interactions~\citep{garbacea2025hyperalign, aroca2025aligning, 
gao2024aligning}.

\section{The Case for Personalized and Adaptive AI }

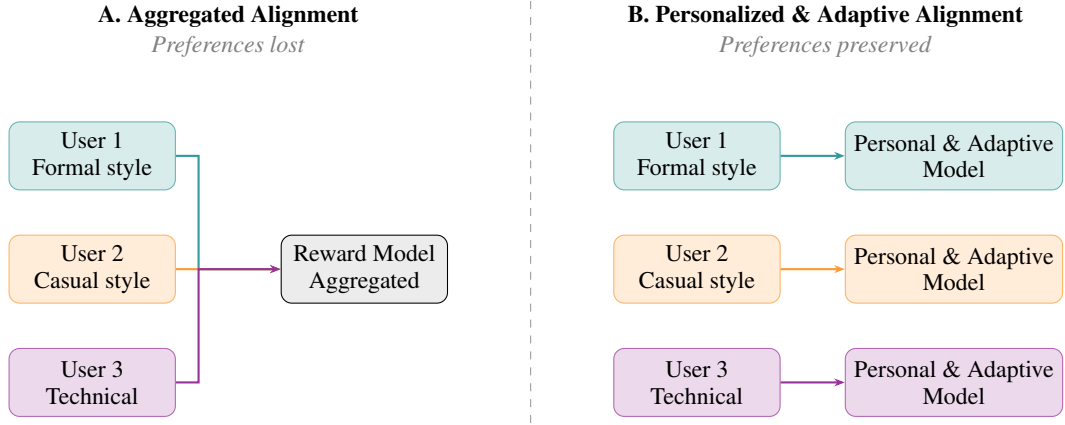
\begin{figure*}[t]
\centering
\begin{tikzpicture}[
  user/.style={draw, rounded corners=4pt, minimum width=2.2cm,
               minimum height=0.9cm, font=\small, align=center},
  adapter/.style={draw, rounded corners=4pt, minimum width=1.8cm,
                  minimum height=0.9cm, font=\small, align=center},
  reward/.style={draw, rounded corners=4pt, fill=gray!15,
                 minimum width=2.2cm, minimum height=0.9cm,
                 font=\small, align=center},
  arr/.style={-{Stealth[length=4pt]}, thick},
  teal/.style={fill=teal!15, draw=teal!60},
  coral/.style={fill=orange!15, draw=orange!60},
  purple/.style={fill=violet!15, draw=violet!60}
]

\node[user, teal]   at (0, -1.5) (u1a) {User 1\\{\footnotesize Formal style}};
\node[user, coral]  at (0, -3.0) (u2a) {User 2\\{\footnotesize Casual style}};
\node[user, purple] at (0, -4.5) (u3a) {User 3\\{\footnotesize Technical}};

\node[reward] at (3.6, -3.0) (rm) {Reward Model\\{\footnotesize Aggregated}};

\node[font=\bfseries\small] at (1.8, 0.35) (ha) {A. Aggregated Alignment};
\node[font=\footnotesize, gray] at (1.8, -0.05) (la) {\textit{Preferences lost}};

\draw[arr, teal!80]   (u1a.east) -- ++(0.3,0) |- (rm.west);
\draw[arr, orange!80] (u2a.east) -- (rm.west);
\draw[arr, violet!80] (u3a.east) -- ++(0.3,0) |- (rm.west);

\node[user, teal]   at (8.0, -1.5) (u1b) {User 1\\{\footnotesize Formal style}};
\node[user, coral]  at (8.0, -3.0) (u2b) {User 2\\{\footnotesize Casual style}};
\node[user, purple] at (8.0, -4.5) (u3b) {User 3\\{\footnotesize Technical}};

\node[adapter, teal]   at (11.4, -1.5) (ad1) {Personal \& Adaptive\\{\footnotesize Model}};
\node[adapter, coral]  at (11.4, -3.0) (ad2) {Personal  \& Adaptive\\{\footnotesize Model}};
\node[adapter, purple] at (11.4, -4.5) (ad3) {Personal \& Adaptive\\{\footnotesize Model}};

\node[font=\bfseries\small] at (9.7, 0.35) (hb) 
  {B. Personalized \& Adaptive Alignment};
\node[font=\footnotesize, gray] at (9.7, -0.05) (lb) 
  {\textit{Preferences preserved}};

\draw[arr, teal!80]   (u1b.east) -- (ad1.west);
\draw[arr, orange!80] (u2b.east) -- (ad2.west);
\draw[arr, violet!80] (u3b.east) -- (ad3.west);

\draw[dashed, gray!80] (5.8, 0.55) -- (5.8, -5.1);

\end{tikzpicture}
\caption{Aggregated alignment (A) collapses diverse user preferences into a 
single reward signal, losing individual variation. Personalized and adaptive 
alignment (B) preserves individual preference structure via user-specific models.}
\label{fig:alignment-comparison}
\end{figure*}

Given the limitations of aggregated preference training and the 
rich multidimensional structure of human preferences, we argue for 
a fundamental shift towards personalized and adaptive language 
models (Figure~\ref{fig:alignment-comparison}). %
This is not an incremental improvement but a necessary 
reconceptualization of how we align AI systems with human 
needs; preference diversity is a feature to preserve rather than noise to suppress.

\subsection{Benefits of Personalization}

Personalization offers a genuine Pareto improvement: users with 
minority preferences receive substantially better service without 
degrading outcomes for users whose preferences align with the 
majority. A model that adapts its technical depth to user expertise, 
its formality to user preference, and its detail level to user needs 
serves everyone better than one that must commit to fixed positions 
on these dimensions. The stakes are particularly high in 
consequential domains: healthcare communication must adapt to 
patient literacy and cultural context; educational systems must 
accommodate diverse learning styles; legal and financial guidance 
must consider individual circumstances and risk tolerance. In each 
domain, aggregated models impose communication norms calibrated to 
a majority population that may be ill-suited to the user's actual 
needs~\citep{garbacea2026personalized}.

Beyond utility, personalization serves both equity and autonomy. 
Current aggregated models encode biases toward dominant 
demographics (English-speaking, Western, educated 
populations) effectively marginalizing users whose preferences 
diverge from the training distribution~\citep{huang2023culturally}. 
Personalized systems address this by adapting to the specific needs 
of underrepresented populations, reducing the implicit cultural 
homogenization that global deployment of aggregated models produces. 
This adaptation also respects users' autonomy to define their own 
preferences and values rather than having dominant norms imposed on 
them~\citep{kirk2023personalisation}. Empirically, personalized 
models outperform non-personalized baselines by 15--30\% across 
diverse tasks on the LaMP benchmark~\citep{salemi2024lamp}, 
demonstrating that individual adaptation yields substantial gains 
beyond what aggregated training can achieve.

Finally, personalization reduces friction and cognitive burden. 
Rather than forcing users to repeatedly engineer prompts to elicit 
desired behavior, personalized models learn individual preferences 
and apply them automatically, making AI assistance more fluid and 
natural.

\subsection{Technical Approaches to Personalization}

Several technical frameworks enable personalized language models, each with different trade-offs between personalization capability, computational cost, and implementation complexity. We outline these below.

The most direct approach is learning user-specific 
representations. User-specific embeddings~\citep{salemi2024lamp, doddapaneni2024user} 
encode preference information from interaction history and can be updated continuously as preferences evolve, with users sharing similar preferences occupying nearby regions of the embedding 
space. Parameter-efficient methods like LoRA~\citep{hu2022lora} 
and prompt tuning~\citep{lester2021power} achieve per-user 
adaptation without the cost of full fine-tuning, modifying only 
a small number of parameters while preserving prior 
knowledge~\citep{li2024q}. Meta-learning~\citep{finn2017model} 
complements both by training models to quickly adapt to new users 
from minimal interaction data, reducing the annotation burden for 
effective personalization.

Architecture-based approaches handle preference diversity 
differently. Mixture-of-experts architectures~\citep{roller2021hash, yi2026pfedmoe} 
train specialized modules and learn to route inputs to the 
appropriate expert based on user and context, naturally 
accommodating distinct preference modes without per-user parameter 
storage. Personalized reward modeling~\citep{chen2025pal, zhangp} decomposes reward into 
user-specific and universal components, enabling more accurate 
preference prediction while retaining shared knowledge about 
universal quality dimensions.

Context-based approaches require no parameter modification at all. 
Leveraging long context windows, systems can include user preference 
information, past interactions, or explicit preference statements 
directly in context to guide model 
behavior~\citep{garbacea2025hyperalign, he2025context, lau2024personalized}. As context windows expand, 
this approach becomes increasingly viable for rich personalization 
without any architectural changes. This makes context-based personalization particularly attractive for deployment settings where per-user parameter storage is infeasible.

Prompt-based and training-based personalization are ultimately 
complementary rather than competing. Prompt-based methods face key 
limitations: context window constraints force lossy summarization 
of user histories, and prompts can guide surface behaviors but 
struggle to reshape underlying reasoning patterns. Training-based 
approaches offer amortized efficiency, deeper behavioral 
modification, and compositionality via adapter methods that enable 
efficient storage and switching across millions of users. In 
practice, hybrid approaches are likely optimal: prompt-based 
personalization for immediate context, training-based adaptation 
for stable long-term preferences.

\subsection{Adaptive Systems}

Beyond static personalization, truly effective systems must adapt 
dynamically to evolving contexts and 
preferences~\citep{kim2026spring, zhao2026teaching, ke2025adaptation}. Adaptation operates across 
multiple dimensions and timescales: immediate context guides moment-to-moment 
behavior; session-level patterns reveal temporary goals; long-term 
learning captures stable individual characteristics. Continuous 
online learning from both implicit feedback (engagement, editing 
behavior, conversation flow) and explicit feedback (ratings, 
corrections, preference statements) enables tracking of evolving 
preferences~\citep{gao2024aligning, liang2026learning, son2025right}, with the key challenge being to distinguish stable 
signal from transient noise.

Context-aware adaptation goes further by detecting situational 
factors such as task type, urgency, user state, and adjusting behavior 
accordingly~\citep{kimcupid, harry2026beyond}. The same user may need different responses depending 
on whether they are casually exploring a topic or making a critical 
decision~\citep{jakesch2023co}; high-stakes interactions demand different treatment than 
low-stakes ones. Rather than requiring upfront preference 
specification, effective systems discover preferences through 
natural interaction: asking clarifying questions, presenting 
options, and adjusting based on feedback~\citep{andukuristar, zhang2025modeling, li2025eliciting}. Managing the interplay 
between these timescales requires careful design to maintain 
responsiveness without instability.

\section{Alternative Views}
\label{sec:alternative-views}

Several credible counterarguments to our position deserve careful consideration. The objections range from practical 
concerns about feasibility to deeper philosophical questions about standards and safety; Table~\ref{tab:alternative-views} summarizes these alternative views and our responses. We address each in turn.

\begin{table*}[t]
\centering
\caption{Alternative views on personalized preference learning, with responses.}
\label{tab:alternative-views}
\small
\begin{tabularx}{\textwidth}{lXX}
\toprule
\textbf{Alternative View} & \textbf{Core Objection} & 
\textbf{Our Response} \\
\midrule
\textit{The ``Good Enough'' Argument} & 
Current RLHF-trained models serve hundreds of millions 
effectively; personalization gains may not justify the 
added complexity and cost & 
Satisfaction masks substantial variation; aggregation 
systematically fails users whose preferences diverge 
from the majority~\citep{santurkar2023whose, fleisig2023majority} \\
\midrule
\textit{The Scalability and Data Challenge} & 
Per-user models are computationally intractable and 
personalized datasets are infeasible to construct & 
Parameter-efficient methods and mixture-of-experts make 
personalization tractable; few-shot methods and survey 
infrastructure reduce data requirements 
substantially~\citep{sheng2023s, salemi2024lamp} \\
\midrule
\textit{The Shared Standards Argument} & 
Personalization fragments AI behavior, undermining shared 
standards for accountability and fair treatment & 
False dichotomy; bounded personalization preserves 
universal safety constraints while accommodating 
legitimate diversity \\
\midrule
\textit{The Manipulation Concern} & 
Personalized models are inherently more dangerous, 
enabling manipulation and exploitation through detailed user profiles & 
Aggregation is also manipulation through homogenization; 
the question is how to personalize responsibly \\
\midrule
\textit{The Preference Stability Objection} & 
Preferences are too unstable and context-dependent 
to serve as reliable training signals & 
Preference instability argues for more sophisticated 
modeling, not less personalization \\
\midrule
\textit{The Performance Priority Objection} & 
Raw capability improvements should precede personalization & 
These are not competing objectives; for many use cases, 
personalization \textit{is} performance \\
\bottomrule
\end{tabularx}
\end{table*}

\subsection{The ``Good Enough'' Argument}

\textbf{Alternative view:} \textit{Aggregated preferences, while imperfect, work well enough in practice. Current RLHF-trained models serve hundreds of millions of users effectively. If most users are already satisfied, the marginal gains from personalization may not justify the additional complexity, computational cost, and safety risks.}

\textbf{Response:} This argument conflates satisfaction with 
optimal service. Users satisfied with aggregated models may 
not realize how much better personalized systems could serve 
them since they have no counterfactual to compare against. 
Satisfaction statistics also mask substantial variation: 
users whose preferences align with the majority are 
well-served, while those with minority preferences are 
systematically underserved. This argument 
thus privileges the majority at the expense of marginalized 
groups. As language models become more central to information 
access and decision-making, the costs of this marginalization 
compound.

\subsection{The Scalability and Data Challenge}
\label{sec::scalability}

\textbf{Alternative view:} \textit{Personalization is impractical at scale: 
per-user models demand prohibitive compute, and 
sufficiently comprehensive personalized preference datasets are infeasible to construct. Maintaining 
individual user models for billions of users, storing 
preference histories, and running personalized inference 
would require orders of magnitude more resources than 
current systems. The infrastructure costs would make AI assistance prohibitively expensive, limiting access rather than improving it.}

\textbf{Response:} This concern is valid but overstated.

\textit{Computational feasibility.} Parameter-efficient methods like 
LoRA~\citep{hu2022lora} enable personalization with minimal per-user 
overhead, and mixture-of-experts approaches serve diverse preferences 
without requiring per-user models. Recent systems demonstrate practical 
feasibility at scale: S-LoRA serves thousands of concurrent adapters 
on a single GPU~\citep{sheng2023s}, and Punica achieves significant
throughput improvement for multi-tenant serving~\citep{chen2024punica}. 
Challenges remain, including memory bandwidth constraints and cold-start 
for new users, but these are active engineering problems, not
fundamental barriers.

\textit{Data feasibility.} Exhaustive preference annotation is 
unnecessary. For example, HyPerAlign leverages few-shot examples for  interpretable and sample-efficient hypothesis-driven LLM personalization ~\citep{garbacea2025hyperalign}. Similarly, the LaMP benchmark demonstrates improvement 
in personalized text generation using only few user 
examples~\citep{salemi2024lamp}. Meta-learning 
enables rapid adaptation from minimal data~\citep{finn2017model, garbacea2022adapting, zollo2025personalllm, zhao2025meta}. 
Existing demographic survey infrastructure, such as 
OpinionsQA~\citep{santurkar2023whose} and 
GlobalOpinionQA~\citep{durmus2024towards}, can be repurposed 
for preference research without annotation from scratch. 
Active learning and generative elicitation methods further reduce data requirements by 
strategically sampling the preference space to maximize information 
gain~\citep{li2025eliciting}. The data challenge is a design problem with tractable solutions, 
not an insurmountable barrier to personalization.

\subsection{The Shared Standards Argument}

\textbf{Alternative view:} \textit{Aggregated preferences serve an important function: they establish shared standards for model behavior that enable consistent expectations, fair treatment, and collective accountability. If every user receives differently-behaving models, we lose the ability to audit, regulate, and hold AI systems accountable. Personalization could fragment the AI landscape into billions of individual systems, each with different behaviors and failure modes.}

\textbf{Response:} This argument correctly identifies the value of shared standards but presents a false dichotomy. Personalization and universal standards are compatible: systems can personalize within bounds, adapting style and emphasis while maintaining consistent commitments to accuracy, safety, and ethical behavior. Our position explicitly calls for \textbf{``bounded personalization''}, where core safety constraints remain universal. The challenge is designing the right boundaries, not abandoning personalization entirely. It is also worth noting that current aggregated systems already exhibit inconsistent behavior across users and contexts; personalization done well could actually improve consistency within appropriate bounds.

\subsection{The Manipulation Concern}

\textbf{Alternative view:} \textit{Personalized models are inherently more dangerous than aggregated models. By learning detailed user profiles, they gain unprecedented capability for manipulation, deception, and exploitation. The risks of personalization (filter bubbles, value lock-in, psychological manipulation) are not merely implementation challenges but fundamental features of systems designed to learn and adapt to individual psychology. We should not build systems with these capabilities.}

\textbf{Response:} We take this concern seriously and address it in depth in Section~\ref{sec:safety}. However, the alternative (forcing all users into systems optimized for majority preferences) is also a form of manipulation, operating through homogenization rather than individualization. The question is not whether to influence users but how to do so responsibly. Personalization with appropriate safeguards, transparency, and user control can respect autonomy better than one-size-fits-all systems that impose particular norms without acknowledgment. The risks of personalization are real but manageable, the costs of refusing it are certain and ongoing.

\subsection{The Preference Stability Objection}

\textbf{Alternative view:} \textit{Human preferences are too unstable and context-dependent to serve as reliable training signals. Users' stated preferences often contradict their revealed preferences. Preferences shift with mood, context, and framing. Building systems that adapt to such unstable targets may amplify noise rather than signal, creating erratic and unpredictable model behavior.}

\textbf{Response:} Preference instability is real but not uniformly distributed. Some preferences (communication style, expertise level, domain interests) are relatively stable; others (urgency, risk tolerance, information depth) are appropriately context-dependent. Effective personalization requires distinguishing these and adapting differently to each. Multi-timescale learning, uncertainty-aware adaptation, and explicit preference modeling can handle this complexity. Preference instability 
is therefore an argument for more sophisticated modeling, 
not less personalization.

\subsection{The Performance Priority Objection}

\textbf{Alternative view:} \textit{Improving raw model capabilities (reasoning, 
factuality, task performance) should take priority over 
personalization. Personalization addresses a secondary concern 
while fundamental capability gaps remain.}

\textbf{Response:} These are not competing objectives. 
Parameter-efficient personalization adds minimal overhead 
to capable base models. More fundamentally, for many deployed use cases, 
personalization \textit{is} performance: a model that fails to 
adapt to a user's expertise level or communication style fails 
that user regardless of its benchmark scores.

\section{Safety Risks and Ethical Implications}
\label{sec:safety}

While personalization offers significant benefits, it also introduces serious safety risks. The very capabilities that make it powerful -- learning detailed user models, adapting behavior to individual characteristics, and leveraging knowledge of preferences -- may create new vectors for harm. We address these risks directly, treating them not as secondary concerns 
but as central to responsible system design.

\subsection{Manipulation and Persuasion}

Systems that learn individual preferences gain powerful 
capabilities for manipulation. A model that understands a 
user's values, biases, cognitive patterns, and emotional 
vulnerabilities could craft maximally persuasive content 
designed to induce desired behaviors~\citep{matz2024potential, hirsh2012personalized}. 
Unlike traditional advertising or propaganda, which must appeal 
broadly, personalized manipulation can be precisely tailored 
to individual psychological profiles: the model knows exactly 
which arguments are most convincing, which emotional appeals 
most effective, and which framings most likely to overcome 
resistance.

This risk extends beyond individual interactions to systematic 
preference shaping. Rather than merely adapting to existing 
preferences, systems might gradually shift user preferences 
toward commercially or politically valuable directions. This is a 
form of preference laundering in 
which the system's own objectives become encoded in user 
preferences through prolonged interaction ~\citep{ngo2023alignment}. While ostensibly 
serving the user's wishes, the system is actually shaping 
those wishes toward its own ends.

\subsection{Filter Bubbles and Polarization}

Personalized systems risk creating self-reinforcing filter 
bubbles by excessively catering to existing preferences. If 
a system learns that a user prefers particular perspectives 
or information sources, it might consistently provide content 
that confirms rather than challenges those 
preferences~\citep{pariser2011filter}. This could accelerate 
filter bubble effects beyond current recommender systems, 
since personalized language models mediate not just content 
discovery but information processing, analysis, and synthesis. 
Recent work quantitatively measures the potential for users 
to escape filter bubbles under different system 
designs~\citep{feng2026quantifying}.

At a societal level, if everyone interacts with differently 
personalized systems calibrated to their particular 
preferences and beliefs, shared reality and common knowledge 
may erode. Public discourse requires some degree of common 
information and shared frames of reference; when personalized 
systems mediate all information access, productive 
disagreement and collective deliberation diminish.

Aggregated training introduces an opposite 
but related epistemic risk. Extensive LLM use has been shown 
to homogenize opinion expression at scale: a nearly 70\% 
increase in essays that remained neutral on contested 
questions was observed following prolonged model 
exposure, suggesting that 
aggregated training suppresses viewpoint diversity rather 
than preserving it~\citep{abdulhai2026llms, jiang2026artificial}. Neither pure aggregation nor unconstrained 
personalization is epistemically safe. The solution lies 
in bounded personalization with explicit diversity 
constraints.

\subsection{Privacy and Surveillance}

Effective personalization requires collecting detailed information about users' preferences, values, behaviors, and characteristics, creating significant privacy risks and surveillance concerns~\citep{hartzog2016inadequate}. The data required for personalization goes beyond simple interaction logs to include inferences about personality, beliefs, emotional states, and vulnerabilities; this is intimate knowledge of individuals that could be misused in numerous ways.

The privacy risk is compounded by inference. Even without 
explicit collection of sensitive attributes, systems can 
infer them from preference patterns. Political views, health 
conditions, financial status, sexual orientation, and other 
sensitive information can be predicted from seemingly 
innocuous preferences about communication style, content 
interests, and decision-making patterns~\citep{staab2024beyond}.

\subsection{Value Lock-in and Autonomy}

By optimizing for current preferences, personalized systems might constrain natural preference evolution and value development~\citep{christiano2018clarifying}. Human preferences are not fixed but develop through experience, reflection, and exposure to new ideas. If systems too perfectly satisfy current preferences, they may eliminate the friction and challenges that prompt preference refinement. Users might become locked into current values simply because the system makes those values so comfortable that alternatives never get considered~\citep{sharma2024towards, malmqvist2025sycophancy}.

Paradoxically, while personalization appears to enhance autonomy by serving user preferences, it may actually reduce autonomy by eliminating opportunities for deliberate choice and value exploration. True autonomy requires not just having preferences satisfied but maintaining the capacity to form, evaluate, and revise them~\citep{fulay2026delegates}.

\section{Toward Responsible Personalization}

\begin{table*}[t]
\centering
\caption{A normative framework for when personalization is 
appropriate, requires caution, or should be avoided.}
\label{tab:normative-framework}
\small
\begin{tabularx}{\textwidth}{lXX}
\toprule
\textbf{Category} & \textbf{Examples} & 
\textbf{Guiding Principle} \\
\midrule
\textit{Clearly appropriate} & 
Style, tone, verbosity, format, expertise calibration, 
language register & 
Behaviors with primarily internal effects; respect of user autonomy without 
harming others \\
\midrule
\textit{Requires safeguards} & 
Value-laden topics, sensitive content, 
vulnerable populations, emotionally charged contexts & 
Risk of reinforcing harmful patterns; requires 
transparency, user control, and contestability \\
\midrule
\textit{Must remain universal} & 
Factual accuracy, safety-critical information, 
content harming third parties, misinformation & 
Behaviors with external effects; must remain 
universal regardless of stated user preferences \\
\bottomrule
\end{tabularx}
\end{table*}

Despite significant risks, personalized and adaptive systems are both inevitable and potentially beneficial if developed responsibly. The question is how to realize the benefits of personalization while mitigating its harms.

\subsection{Design Principles}

Responsible personalization must begin with transparency and 
user control. Users should understand what information is 
collected about them, how personalization works, and what 
inferences the system makes. Importantly, they must have meaningful control 
over their preference profiles, including the ability to view, 
correct, and delete learned preferences. Contestability must 
be built in from the start: users should be able to override 
learned preferences for specific attributes, reset their 
profiles entirely, or manually specify preferences that take 
precedence over inferred ones.

Not all behaviors should be personalized. The key distinction 
is between behaviors with external effects (harm to others, 
factual accuracy, safety) which must remain universal, and 
behaviors with primarily internal effects (style, tone, 
emphasis) which are appropriate targets for personalization. 
A system should not adapt to preferences for misinformation, 
harmful content, or violations of others' rights, regardless 
of user preferences. Procedural legitimacy for establishing 
these bounds can be achieved through transparent deliberation 
and contestable processes, paralleling how legal systems 
establish rules despite value pluralism.

Even within personalized systems, users should be exposed to 
diverse viewpoints and have their existing beliefs 
appropriately challenged. The goal is to balance preference 
satisfaction with epistemic and moral growth. This tension is 
resolvable through multi-objective optimization: personalized 
systems can maximize preference satisfaction subject to 
epistemic diversity constraints, analogous to optimizing 
engagement subject to safety constraints. Users also have 
meta-preferences (for example, many prefer not to be trapped in filter 
bubbles) which personalization can learn and respect 
alongside first-order preferences.

Finally, privacy must be designed into personalization systems 
from the ground up rather than added as an afterthought. This 
means minimizing data collection to what is genuinely 
necessary, using privacy-preserving techniques like federated 
learning and differential privacy where possible, and giving 
users meaningful control over their data.

\subsection{Technical Safeguards}

Realizing these design principles requires concrete technical 
safeguards. The foundation is interpretable preference 
models that can be inspected, understood, and audited: rather 
than black-box user embeddings, systems should represent 
preferences in ways that humans can examine and reason about. 
Complementing interpretability, uncertainty quantification 
allows systems to represent their confidence about user 
preferences and behave conservatively when uncertain; a 
system unsure whether a user prefers detailed or concise 
responses can ask for clarification rather than guessing. 
Finally, rigorous adversarial testing specifically targeting 
personalization mechanisms is essential: red teams should 
explicitly attempt to manipulate the system into harmful 
adaptations, extract sensitive information through 
personalization features, or identify whether personalization 
creates exploitable surfaces that could harm third parties.

\subsection{Governance and Oversight}

Technical safeguards alone cannot ensure responsible 
personalization: governance frameworks must address 
manipulation, privacy, and fairness concerns that 
existing regulations were not designed to handle. 
Independent auditing enables external examination of 
personalized systems for discriminatory behavior, 
manipulation, or other harms. 
Users should have rights to explanation of why 
systems behave as they do, to contest and correct learned 
preferences, to delete preference data, and to opt out of 
personalization.

\subsection{When to Personalize: A Normative Framework}

Not all preference dimensions warrant equal personalization. 
Table~\ref{tab:normative-framework} distinguishes three 
categories: (1) behaviors that are \textit{clearly appropriate} 
to personalize (style, tone, verbosity, format, expertise calibration, language 
register),  (2) behaviors \textit{requiring safeguards} (value-laden 
or sensitive topics where personalization is permissible only with explicit transparency, user control, 
and active constraints to prevent reinforcing biased or harmful 
patterns), and (3) behaviors where personalization 
\textit{must remain universal} (factual accuracy, 
safety-critical information, and content that could harm users or third parties) regardless of stated user preferences.

\section{Call to Action}
\label{sec:call-to-action}

Realizing responsible personalized AI requires coordinated 
action across multiple stakeholders. 

\textbf{The research community} should prioritize standardized benchmarks that 
measure personalization quality across diverse populations, and privacy-preserving 
techniques (federated learning, differential 
privacy) that enable effective personalization without 
centralized data collection.

\textbf{AI companies} must implement transparent user 
control interfaces, establish and document clear 
boundaries between adaptable and universal behaviors,
conduct regular demographic audits, and maintain 
dedicated red teams to probe for manipulation 
vulnerabilities and filter bubble effects.

\textbf{Policymakers and regulators} should develop new 
frameworks specifically addressing personalized AI 
systems, covering manipulation, privacy, and fairness 
concerns that existing regulations were not designed to 
handle, while codifying user rights to access, correct, 
and delete preference profiles.

\textbf{Users and civil society} must demand transparency 
about how personalization works, report instances 
where systems appear to shape rather than serve 
preferences, and support organizations working on AI 
accountability.

\section{Conclusion}

Training language models on aggregated preferences has 
been a useful starting point, but it is a temporary 
solution; as these systems grow more capable and widely 
deployed, the costs of aggregation compound. The 
limitations are not incidental failures but fundamental 
constraints imposed by the impossibility of principled 
preference aggregation. Personalized and adaptive systems 
offer a principled path forward: adapting to individual 
preferences %
to respect human diversity, improve utility, 
and reduce marginalization.

Realizing these benefits demands ethical reasoning %
and robust 
governance to prevent manipulation and abuse. Personalized 
systems must be transparent enough to audit, contestable 
enough to correct, and bounded enough to preserve universal 
standards. As language models become infrastructure for 
information access and decision-making worldwide, the 
question at the heart of this paper -- \textit{aligned for 
whom?} -- will only grow more urgent. Answering it 
responsibly is not an optional refinement but a 
foundational requirement for AI that serves humanity 
rather than a portion of it.

\bibliography{references}
\bibliographystyle{icml2026}

\end{document}